\documentclass{article}
\usepackage{graphicx} 
\usepackage{amsmath}
\usepackage{amssymb}
\usepackage{arydshln}
\usepackage{multirow}
\usepackage{caption}
\usepackage{wrapfig}
\usepackage{tcolorbox}
\usepackage{enumitem}
\usepackage{subcaption}

\usepackage[nonatbib, preprint]{neurips_2024}

\usepackage[utf8]{inputenc} %
\usepackage[T1]{fontenc}    %
\usepackage{hyperref}       %
\usepackage{url}            %
\usepackage{booktabs}       %
\usepackage{amsfonts}       %
\usepackage{nicefrac}       %
\usepackage{microtype}      %
\usepackage{xcolor}         %
\usepackage{makecell}

\usepackage{amsmath,amsfonts,amssymb}
\usepackage{xspace}

\providecommand{\eg}{\textit{e.g.}\@\xspace}
\providecommand{\ie}{\textit{i.e.}\@\xspace}

\let\oldtitle\title
\renewcommand{\title}[1]{\oldtitle{\large #1}}

\title{MUSAR: Exploring \underline{Mu}lti-\underline{S}ubject Customization from Single-Subject Dataset via \underline{A}ttention \underline{R}outing}

\author{%
Zinan Guo \quad
Pengze Zhang \quad
Yanze Wu\thanks{Corresponding author} \quad
Chong Mou \quad
Songtao Zhao \quad
Qian He\\
Bytedance Intelligent Creation \\
}

\begin{document}

\maketitle
\begin{figure}[ht]
  \centering
  \includegraphics[width=0.95\textwidth]{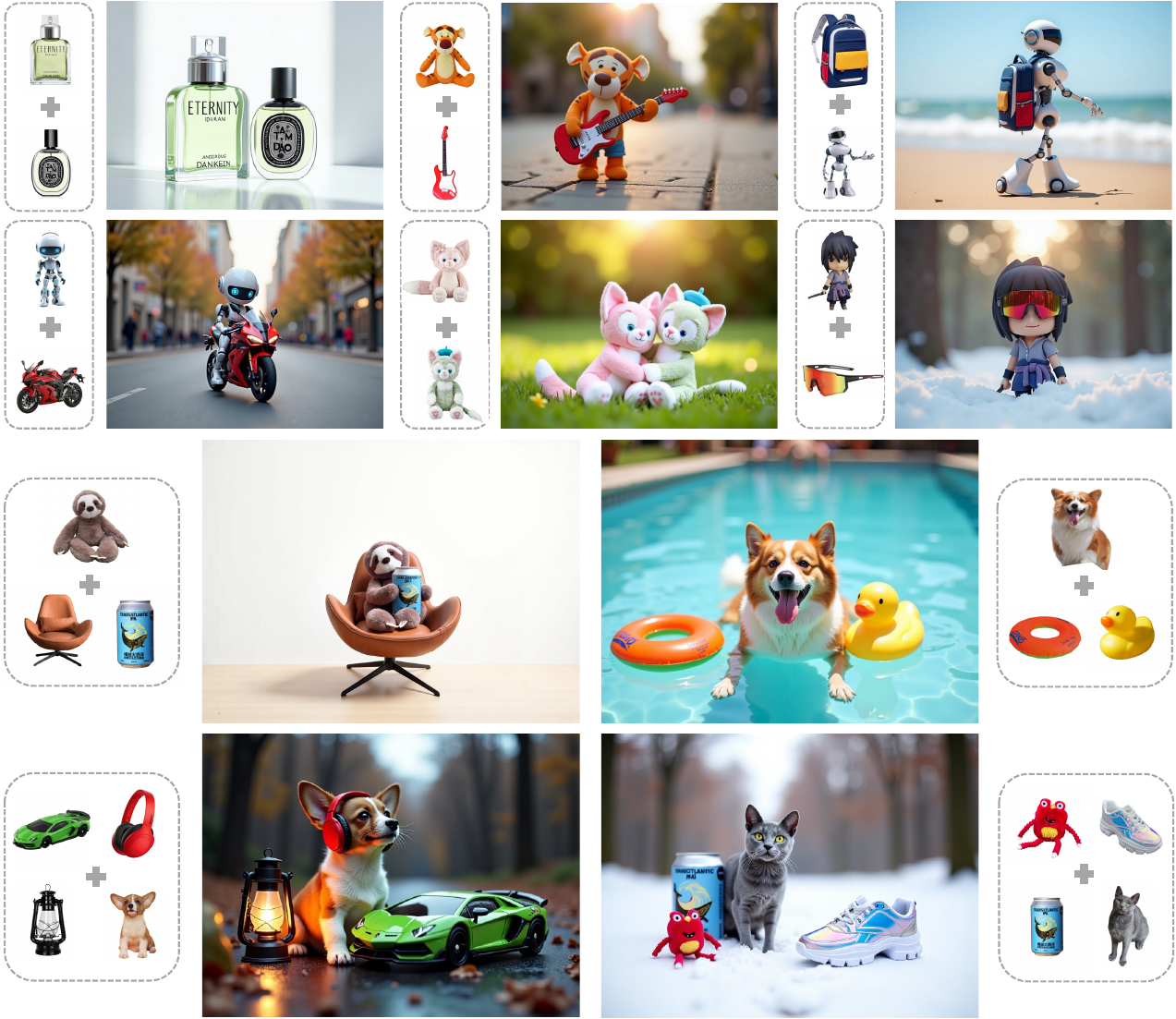}
  \caption{Breaking the data barrier, MUSAR enables remarkable multi-subject customization from solely single-subject dataset, demonstrating scalable generalization as the number of subjects grows.}
  \label{fig:teaser}
\end{figure}

\begin{abstract}\label{sec:abstract}

Current multi-subject customization approaches encounter two critical challenges: the difficulty in acquiring diverse multi-subject training data, and attribute entanglement across different subjects.
To bridge these gaps, we propose MUSAR - a simple yet effective framework to achieve robust multi-subject customization while requiring only single-subject training data.
Firstly, to break the data limitation, we introduce debiased diptych learning.
It constructs diptych training pairs from single-subject images to facilitate multi-subject learning, while actively correcting the distribution bias introduced by diptych construction via static attention routing and dual-branch LoRA.
Secondly, to eliminate cross-subject entanglement, we introduce dynamic attention routing mechanism, which adaptively establishes bijective mappings between generated images and conditional subjects.
This design not only achieves decoupling of multi-subject representations but also maintains scalable generalization performance with increasing reference subjects.
Comprehensive experiments demonstrate that our MUSAR outperforms existing methods - even those trained on multi-subject dataset - in image quality, subject consistency, and interaction naturalness, despite requiring only single-subject dataset.

\end{abstract}

\section{Introduction}\label{sec:intro}
Customized text-to-image (T2I) generation is the process of creating images from text prompts with additional user-specific inputs, such as personal identity, subject, or style, to produce results tailored to individual needs. It is widely applied in areas such as personalized content creation, virtual try-on, creative design, and marketing. To circumvent the need for test-time fine-tuning~\cite{gal2023textual,ruiz2023dreambooth,hu2022lora} on each user input, developing tuning-free approaches for image customization has emerged as a central focus of current research.

For T2I diffusion models that utilize UNet as the backbone~\cite{dhariwal2021diffusion,rombach2022sd,podell2024sdxl}, mainstream image customization methods~\cite{ye2023ipadapter,wei2023elite,chen2024anydoor,zhang2024ssr,guo2024pulid,wang2025ms,ma2024subject} typically involve using an encoder (\eg, CLIP~\cite{radford2021clip}) to extract features from the reference image, which are then injected into the model—alongside the text—at the cross-attention layers of the UNet. Despite their progress, these methods still face challenges such as insufficient fidelity—due to information loss during feature extraction—and the need to meticulously design task-specific adapters for each application.

With the emergence of diffusion transformers (DiT)~\cite{peebles2023dit,esser2024sd3,flux2024}, a new paradigm for controllable generation has recently gained traction~\cite{tan2024ominicontrol,cai2024diffusion,mao2025ace,le2024one}. These methods employ DiT's pre-trained VAE to extract features from the reference image, which are concatenated with the text modality and the noisy image modality along the sequence dimension. The integrated sequence is then processed by the multi-modal self-attention block (\ie, MMDiT~\cite{esser2024sd3} block) in the DiT, enabling information interaction and  conditional fusion. Owing to its simple and unified design that can be applied across various tasks~\cite{le2024one}, as well as the substantial reduction in information loss achieved through the use of VAE, this class of methods exhibits clear advantages over previous approaches. However, despite achieving notable improvements in single-subject customization~\cite{tan2024ominicontrol}, extending these methods to multi-subject scenarios remains a significant challenge. Firstly, the vast majority of previous works~\cite{ma2024subject,zhang2024ssr,wang2025ms} on multi-subject customization are tailored for UNet architecture and cannot be directly transferred to the DiT model. Secondly, the few existing multi-subject customization methods~\cite{chen2024unireal} built on the unified DiT framework~\cite{chen2024unireal} heavily rely on large-scale multi-subject paired datasets, which are often difficult to construct or collect. Lastly, in the absence of strong model priors (\eg, fine-tuned from video diffusion model~\cite{chen2024unireal}), distinguishing the features of different reference images and avoiding attribute entanglement remain non-trivial challenges.

In this study, we propose MUSAR to address two key challenges in multi-subject text-to-image (T2I) generation: heavy reliance on large-scale multi-subject datasets and attribute entanglement across subjects.
First, we discover that diptych-based training—constructed from single-subject data effectively handles multi-subject generation while mitigating data scarcity. 
However, naive diptych training leads to mode collapse due to the inherent bias in learning diptych pairs.
To resolve this, we introduce de-biased diptych learning, incorporating two strategies, i.e., static attention routing and dual-branch Lora mechanisms, to alleviate the systematic bias introduced by diptych data.
Second, we observe that in T2I generation, distinct subjects (e.g., "Einstein and Newton shaking hands") are typically rendered without identity confusion by state-of-the-art models~\cite{podell2024sdxl,flux2024}, implying that each noisy image token can be mapped to its corresponding subject in the prompt. 
Leveraging this insight, we propose dynamic attention routing mechanism, constraining each noisy token to attend only to reference tokens of its associated subject during self-attention, significantly alleviating attribute entanglement.
Thanks to our carefully designed framework, our MUSAR achieves robust generalization to multi-subject generation tasks while requiring only single-image training data (Figure \ref{fig:teaser}), even outperforming existing approaches that rely on large-scale multi-subject datasets.

We summarize the contributions as follows. (1) We circumvent the difficulty of acquiring high-quality multi-subject data by training solely on diptych data constructed from concatenated single-subject samples, and further mitigate potential diptych-induced biases through static attention routing and dual-branch LoRA mechanism. (2) We propose a dynamic attention routing mechanism that adaptively aligns image regions with their corresponding condition subjects, effectively preventing cross-subject entanglement while maintaining scalability for  increasing reference subjects. (3) Experimental results demonstrate that our method enables flexible and coherent multi-subject interactions while maintaining high fidelity using only single-object datasets.

\section{Related Work}\label{sec:related}

\subsection{Diffusion models}
Diffusion models ~\cite{pmlr-v37-sohl-dickstein15, NEURIPS2020_4c5bcfec, NEURIPS2019_3001ef25, dhariwal2021diffusion, pmlr-v202-song23a, NEURIPS2022_a98846e9, lipman2023flow} have emerged as a cornerstone of generative tasks, particularly in text-to-image synthesis. Early UNet-based designs ~\cite{rombach2022sd, podell2024sdxl} integrated text conditioning via cross-attention within convolutional backbones. These works laid the groundwork for diffusion models, enabling a wide range of downstream tasks such as customized text-to-image generation~\cite{zhang2023adding, mou2024t2i, ye2023ipadapter}, image inpainting~\cite{Lugmayr_2022_CVPR,Xie_2023_CVPR}, image-to-image translation ~\cite{10.1145/3528233.3530757, NEURIPS2022_177d68f4}, and image editing ~\cite{Avrahami_2022_CVPR, meng2022sdedit}. Recent transformer-based models such as the diffusion transformer (DiT) ~\cite{peebles2023dit,esser2024sd3,flux2024} represent a significant advance by incorporating full attention to simultaneously model both intra-image and text-image interactions. This architectural has proven to be highly efficient and has brought substantial enhancements in downstream tasks, as evidenced by \cite{lhhuang2024iclora, lhhuang2024groupdiffusion,avrahami2024stableflow, tan2024ominicontrol}. However, it unavoidably leads to feature entanglement when dealing with multi-condition inputs. This entanglement presents formidable challenges for applications that require fine-grained control, such as multi-subject customization.

\subsection{Subject Customization}
Initial research on subject customization mainly employed in UNet-based diffusion models, which can be broadly classified into two paradigms: test-time fine-tuning and fine-tuning-free paradigms. The test-time fine-tuning methods, exemplified by works like ~\cite{voynov2023p+,Ruiz_2023_CVPR,Kumari_2023_CVPR,Han_2023_ICCV,Gal_2023_TOG,gal2023an}, typically involve refining textual embeddings or adjusting model parameters to achieve subject-specific adaptation. Although effective, these approaches require computationally expensive optimization for each new subject, significantly limiting their practical applicability. To overcome this constraint, researchers developed tuning-free alternatives ~\cite{ye2023ipadapter, wei2023elite, zhang2024ssr, chen2024anydoor, guo2024pulid, wang2025ms, ma2024subject, Huang_2024_CVPR} that employ external encoders to represent target subjects, subsequently injecting these representations into pre-trained models via lightweight adapter modules. Some extensions of these methods ~\cite{ma2024subject,zhang2024ssr,wang2025ms} have demonstrated potential for multi-subject generation tasks. Nevertheless, their performance remains constrained by inherent limitations of the U-Net based diffusion model and the absence of specialized designs for handling multiple subjects.

The emergence of diffusion transformers (DiTs) ~\cite{dhariwal2021diffusion,rombach2022sd,podell2024sdxl} has significantly transformed the paradigm of subject customization. Unlike traditional U-Net-based approaches, most DiT-based methods ~\cite{tan2024ominicontrol,cai2024diffusion,mao2025ace,le2024one} adopt a unified conditioning strategy that jointly processes text embeddings, latent tokens, and VAE-encoded condition subjects. This unified design enables for more effective interaction via full attention mechanisms, demonstrating superior performance in single-subject generation tasks. Nevertheless, this architecture faces inherent limitations in multi-object generation, as the global attention mechanism induces feature interference between objects, resulting in attribute entanglement and identity degradation. Recent work \cite{chen2024unireal} has attempted to address this by constructing video-derived paired training data. However, they not only require massive amounts of carefully aligned paired data, but lacks solutions to prevent attribute entanglement. Consequently, achieving robust multi-subject customization within DiT frameworks remains an open research challenge.

\section{Methods}\label{sec:methods}
Our MUSAR framework addresses multi-subject customization through two key components.
Firstly, to overcome the scarcity of multi-subject datasets, we propose \textbf{De-biased Diptych Learning} strategy. 
It simultaneously enhances multi-object preservation through diptych data construction, and reduces learning bias via our Static Attention Routing and Dual-branch LoRA techniques. 
Secondly, we propose \textbf{Dynamic Attention Routing} to address the critical issue of subject entanglement. 
It dynamically establishes bijective mappings between image regions and condition subjects, effectively eliminating cross-object interference through selective attention masking of non-corresponding conditions.
The following sections provide detailed descriptions of these core components.

\subsection{Preliminary}
The Diffusion Transformer (DiT) represents a state-of-the-art framework for image generation by jointly processing noisy image tokens ${X}\in \mathbb{R}^{n \times d}$ and prompt tokens $T\in \mathbb{R}^{m \times d}$  through a unified multi-modal attention mechanism, where $d$ corresponds to the latent dimension, while $m$ and $n$ correspond to the sequence lengths of text and image tokens respectively. 
FLUX.1, as a DiT implementation, employs a specialized architecture combining adaptive layer normalization modules with Multi-Modal Attention (MMA) blocks. 
Within this framework, both image and text tokens are linearly projected into query ($Q$), key ($K$), and value ($V$) representations, enabling cross-modal attention across all tokens:
\begin{equation}
    \text{MMA}([T;X]) = \text{softmax}\left(\frac{{Q}{K}^{\top}}{\sqrt{d}}
    + M\right)V, \label{eq:mma}
\end{equation}
where $[T; X]$ denotes the concatenation of noise image and prompt tokens. 
$M\in \mathbb{R}^{(l) \times (l)}$($l = m+ n$) serves as the attention flow matrix that regulates cross-modal interactions. 
Each entry $M_{i,j}$ controls the attention strength between token pairs, where $M_{i,j} = 0$ permits full cross-attention between image token $i$ and $j$, while $M_{i,j} = -\infty$ completely blocks their interaction. 
In FLUX.1, $M$ is initialized as a zero matrix, enabling unconstrained bidirectional attention across all image-text tokens.

\begin{figure}[!t]
    \centering
    \includegraphics[width=1.0\textwidth]{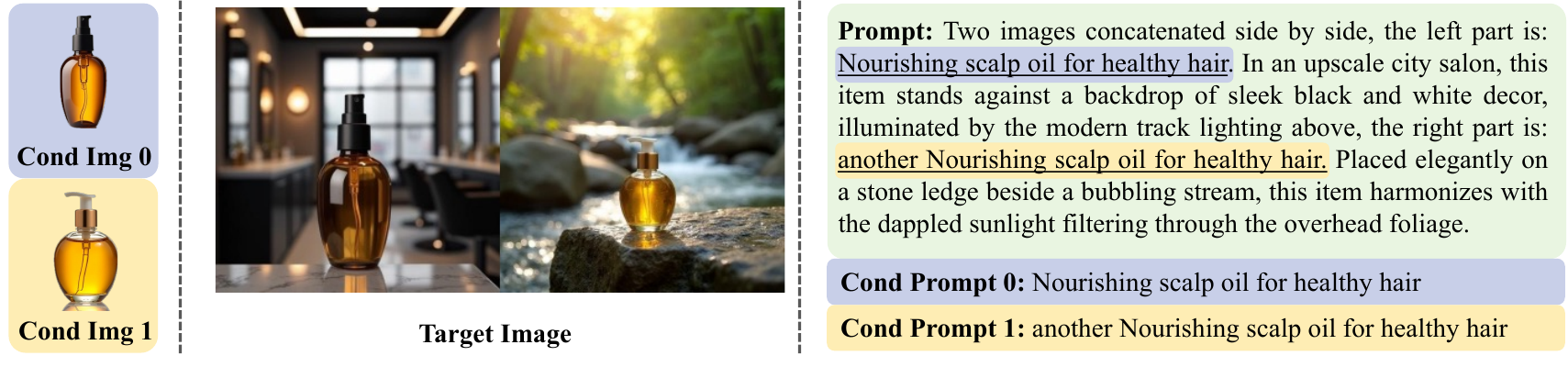}
    \caption{A sample of diptych learning. We pairing existing single-subject data, creating multi-condition and prompts inputs, and diptych targets for effective multi-subject learning.}
    \label{fig:dataset}
\end{figure}
\subsection{De-biased Diptych Learning}
\subsubsection{Diptych Learning}
Models trained on single-subject datasets are struggling in multi-subject customization. 
Meanwhile, the collection of multi-subject datasets encounters formidable obstacles in terms of both data construction and annotation.
To bridge this critical gap, as shown in Figure ~\ref{fig:dataset},  we present a simple yet effective framework to construct multi-subject datasets based on existing single-subject datasets. 
For data construction, we randomly select pairs of distinct subjects from single-subject datasets as conditional references, then concatenate their target images as the diptych target image. 
For prompt engineering, conditional prompts are derived directly from original single-subject descriptions, with the addition of the "another" modifier to disambiguate identical subject types; target image prompts are generated using a structured two-column template, where conditional prompts are adaptively inserted into corresponding left/right column descriptions. 
This paradigm successfully emulates authentic multi-subject scenarios while preserving crucial visual-textual relationships.

With the constructed diptych data, we extract representation for each condition $i$: condition image tokens ${CI}^i \in \mathbb{R}^{n' \times d}$ extracted by VAE, and condition prompt tokens ${CT}^i\in \mathbb{R}^{m' \times d}$ by text encoder.
Then these tokens are directly concatenated with the text tokens $T$ and the noisy image tokens $X$ to form the composite input $[{CI}^1; {CT}^1; ...; {CI}^c; {CT}^c; T; X] \in \mathbb{R}^{(c \times l'+l)}$ for the DiT module, where $l' = m' + n'$ denotes the combined token length of text and image of conditions $i$, and $c$ is the number of condition subjects.
This formulation naturally extends the attention flow matrix $M$ (Eq. ~\ref{eq:mma}) to dimensions$\mathbb{R}^{(c \times l'+l) \times (c \times l'+l)}$, enabling the model to learn coherent multi-subject preservation on the target diptych images.
It is worth noting that $c$ is set to 2 during training due to the diptych data, though the model supports arbitrary numbers of conditions at inference.

\begin{figure}[!t]
    \centering
    \includegraphics[width=1.0\textwidth]{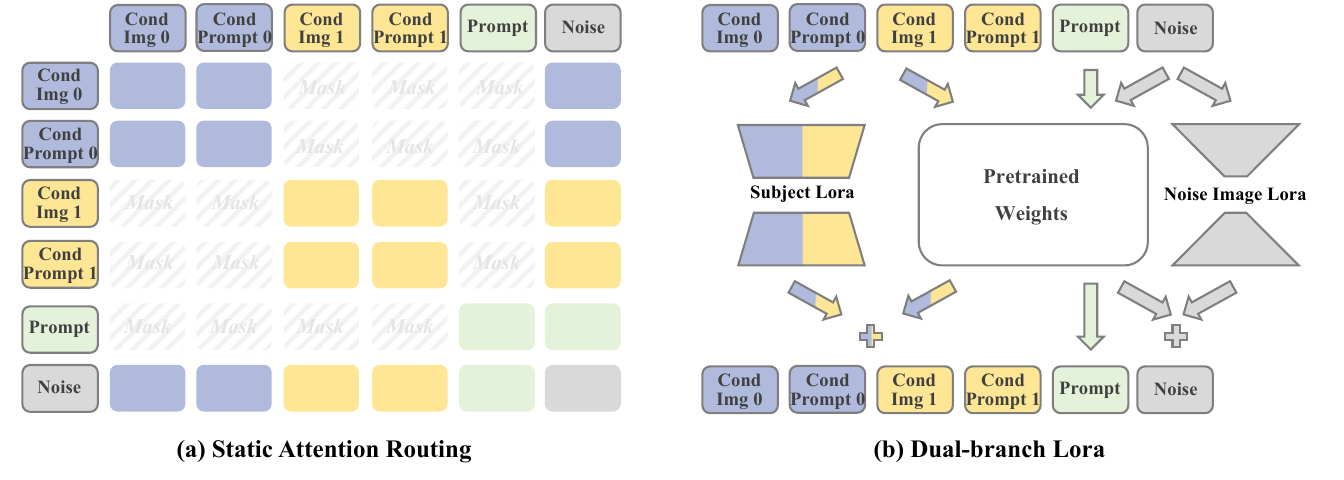}
    \caption{Two strategies to mitigate  learning diptych biases. (a) Static Attention Routing: a routing mechanism that prevents prompt-condition contamination and inter-condition interactions. (b) Dual-branch LoRA: specific LoRA pathways are selectively activated based on input condition types.}
    \label{fig:sar}
\end{figure}

\subsubsection{Diptych Biases Mitigation}
Although the straightforward diptych data construction effectively enables multi-subject learning, it inadvertently introduces two distinct diptych-induced biases: (1) prompt bias, where the use of diptych templates in prompt input corrupts the pre-trained model's text-to-image generation priors, and (2) layout bias, where the uniform diptych arrangement causes the model to develop a strong inherent preference for this specific spatial pattern. 
To mitigate these biases, we introduce two key strategies: static attention routing and dual-branch LoRA.

\textbf{Static Attention Routing.}
To effectively address diptych bias induced by input prompts, we propose a static feature routing mechanism that operates across multiple conditions (Figure~\ref{fig:sar}(a)), comprising two key components: (1) Prompt-Condition Decoupling. The attention flow between the prompt and all conditional features is disabled, i.e., $M_{c\times l':c\times l'+m, 0:c\times l'}=M_{0:c\times l', c\times l':c\times l'+m}=-\infty$. This effectively blocks the transmission of prompt-induced bias to the conditional learning pathway, ensuring the model focuses on preserving individual object characteristics. (2) Inter-Condition Isolation. We further enforce strict separation between different conditional inputs, i.e., $M_{i\times l':(i+1)l', j\times l':(j+1)l'} = M_{j\times l':(j+1)l', i\times l':(i+1)l'}=-\infty$, where $0 \leq i,j \leq c - 1$ and $i \neq j$.
This design minimizes cross-condition interference while enhancing feature discriminability, significantly improving the model's multi-subject generation robustness.

\textbf{Dual-branch LoRA.}
To effectively mitigate diptych bias while learning the multi-subject generation, we propose a specialized LoRA ~\cite{hu2022lora} optimization strategy that differentially processes each input (Figure~\ref{fig:sar} (b)): (1) For prompt input, we intentionally freeze the corresponding weights without LoRA fine-tuning, since the exhibiting inherent template bias that could reinforce two-column image priors. (2) For noise image and conditional inputs, we design a dual-branch LoRA, incorporating a low-rank noisy image LoRA to suppress layout bias learning from the target image, and a high-rank subject LoRA to efficiently learn multi-subject preservation. This asymmetric rank design creates balanced feature learning that simultaneously suppresses spatial overfitting while enhancing-subject representation learning.

\subsection{Dynamic Attention Routing}
\begin{figure}[!t]
    \centering
    \includegraphics[width=1.0\textwidth]{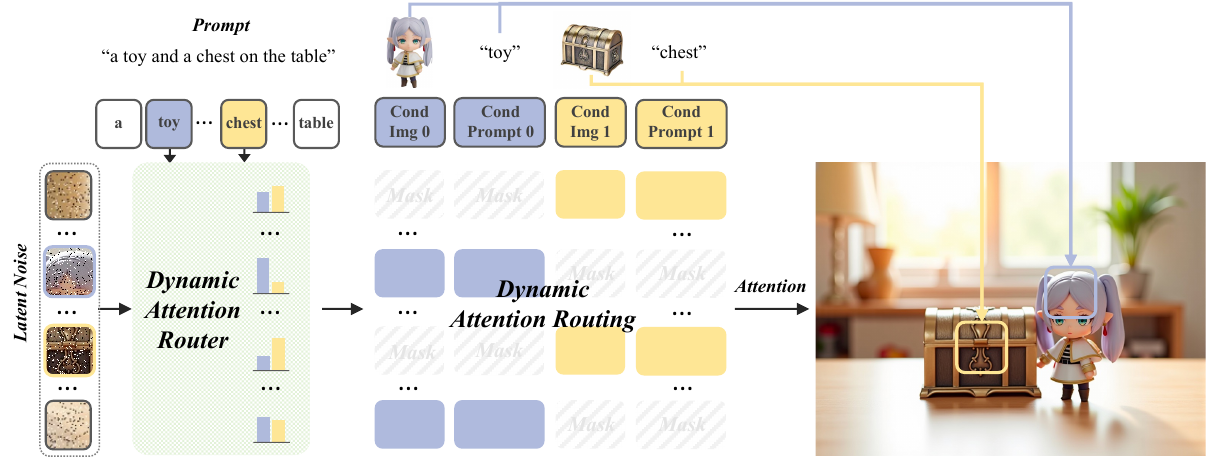}
    \caption{Dynamic Attention Routing enforces a bijective mapping between noise tokens and condition subjects, effectively mitigating multi-subject feature entanglement.}
    \label{fig:dar}
\end{figure}
Although diptych learning and static attention routing reduce inter-subject feature entanglement, significant entanglement remains for semantically similar subjects (e.g., subjects of the same category).
We believe that this phenomenon originates from the attention flow between noise tokens and condition tokens.
Such flow enables the noise tokens to indiscriminately attend to and blend features across multiple conditions, leading to spatial superposition of conflicting object features.
A straightforward solution~\cite{Huang_2024_CVPR} is to predict subject-specific masks at inference time, enforcing the spatial separation of feature injection through constrained attention regions.
However, these approaches present two fundamental challenges: (1) the mask predicted from intermediate text-to-image timestep often shows significant shape discrepancies with target subjects, leading to irreversible information loss, and (2) they typically require extensive tuning of optimal timestep and specific network blocks for mask prediction, severely hindering cross-architecture generalization.
To resolve these compounded challenges, we propose Dynamic Attention Routing that adaptively determines optimal injection subject targets for each noise token.

The architecture of Dynamic Attention Routing is illustrated in Figure \ref{fig:dar}. The routing process begins with computing a similarity matrix $S\in \mathbb{R}^{n \times m}$ between noise tokens and prompt tokens:
\begin{equation}
    S = \text{softmax}\left(\frac{Q_X K^{\top}_T}{\sqrt{d}}
    \right), \label{eq:mma}
\end{equation}
where $Q_X \in \mathbb{R}^{n \times d}$ and $K_T \in \mathbb{R}^{m \times d}$ denote the projected query and key matrices derived from noise tokens $X$ and prompt tokens $T$ respectively.
As discussed in the introduction, modern text-to-image models can correctly map noisy image tokens to their corresponding textual subjects. 
Building on this insight, we establish condition-level associations by computing noise-condition affinity scores, obtained through averaging across each condition’s relevant tokens.
Let $p_k$ denotes the starting index and $l_k$ specifies the length of the token subsequence $T_{p_k:p_k+l_k}$ corresponding to the $k$-th condition in the prompt tokens.
The affinity score between noise token $i$ and condition $j$ can be expressed as:
\begin{equation}
    S^{*}_{i,k} = \frac{1}{l_{k}} \sum_{z=0}^{l_{k} - 1} S_{i,p_k+z},
    \quad  k \in \{0, \ldots, c-1\},
\end{equation} 
yielding the noise-condition affinity matrix $S^{*}\in \mathbb{R}^{n \times c}$. 
Based on this affinity measure, we assign each noise token to its maximally relevant condition by taking the argmax function over $S^{*}$, while simultaneously masking attention to all other competing conditions:
\begin{equation}
    \text{M}_{i,j}=
    \begin{cases} 
    -\infty & \text{if } \left\lfloor \frac{j}{l'} \right\rfloor \neq \underset{k \in \{0, ..., c-1\}}{\arg\max} \left( S^{*}_{i, k} \right) \\
    0 & \text{if } \left\lfloor \frac{j}{l'} \right\rfloor = \underset{k \in \{0, ..., c-1\}}{\arg\max} \left( S^{*}_{i, k} \right)
    \end{cases},
    \quad
    \text{for }
    \begin{cases} 
        c \times l' + m \leq i < c \times l'+l \\
        0 \leq j < c \times l'
    \end{cases}.
\end{equation}
In this way, as shown in the right part of the Figure \ref{fig:dar}, the proposed dynamic attention routing enforces a strict bijective mappings between noise tokens and condition subjects, effectively eliminating feature entanglement. 
It is worth noting that while the Dynamic Attention Routing assigns a condition for all all noise tokens - including background, the inherently low correlation between background and condition tokens renders this mandatory selection negligible.

\begin{figure}[!t]
    \centering
    \includegraphics[width=1.0\textwidth]{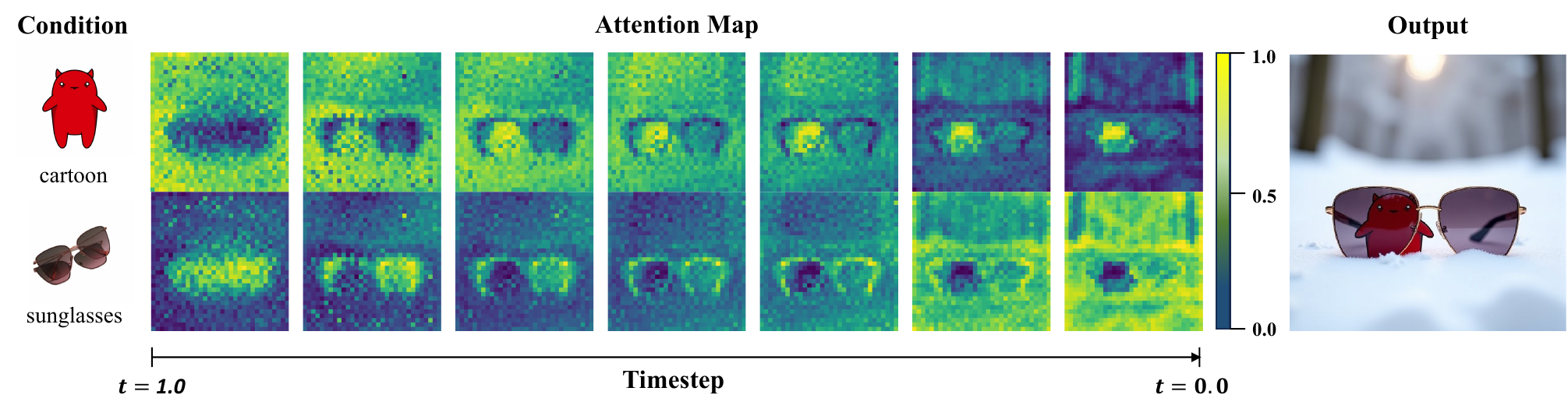}
    \caption{Visualization of the noise-condition affinity score $S^{*}$ in Dynamic Attention Routing. Each row displays dynamic routing probabilities per condition, demonstrating how adaptive attention selectively focuses on different conditions throughout the denoising process.}
    \label{fig:classification}
\end{figure}

We further visualize the noise-conditional affinity matrix $S^{*}$ in Figure \ref{fig:classification} to verify the effectiveness of the proposed dynamic attention routing.
It can be seen that (1) $S^{*}$ can successfully establish clear relation between noise and condition tokens throughout the diffusion process, even at large timesteps ($t \rightarrow 1$);
(2) each noise token adaptively switches its focus condition across timesteps, enabling more natural multi-object interactions (e.g., perspective relationships between sunglasses and cartoon characters).

\section{Experiments}\label{sec:exp}

\subsection{Implementation Details}

We build our model based on FLUX.1-dev~\cite{flux2024} and fine-tune it with Dual-branch LoRA. 
Specifically, we use a LoRA rank of 128 for subject LoRA, and 4 for the noisy image LoRA. 
We construct our training set from  Subject200K~\cite{tan2024ominicontrol}, retaining only samples with the maximum quality rating ($\text{score}=5$), resulting in 111,761 high-fidelity single-subject paired samples.

The training process is divided into three stages, progressing from easy to hard. 
The initial stage (20,000 iterations) establishes fundamental capabilities through exclusive single-subject training, developing robust subject-specific adaptation.
Building upon this foundation, the second stage (10,000 iterations) implements a strategic mixed regime combining 80\% randomly paired diptych data with 20\% single-subject samples, cultivating essential cross-subject discriminative abilities for effective multi-subject customization. 
This enables the model to develop robust cross-subject discriminative capabilities essential for handling multi-subject customization.
The final stage (10,000 iterations) replaces random pairings with same-category diptych constructions, compelling the model to master fine-grained intra-class distinctions while mitigating attribute entanglement.
All experiments were conducted on 8 NVIDIA A100 GPUs, with a batch size of 8, a learning rate of 1e-5, and a training resolution of $512 \times 512$.

We evaluate our method's performance across both single-subject and multi-subject customization tasks. 
For single-subject evaluation, we employ the complete set of 750 test samples from the DreamBench dataset~\cite{ruiz2023dreambooth}. 
For multi-subject scenarios, we construct test cases by pairing subjects from DreamBench to create 60 unique pairs, along with 20 composed triplets, resulting in a comprehensive set of 80 multi-subject test samples.
Following previous works, we measured the model's quantitative performance through image and text fidelity. 
For image fidelity, we used the CLIP~\cite{radford2021learningtransferablevisualmodels} and DINO~\cite{caron2021emergingpropertiesselfsupervisedvision} models to calculate the cosine similarity between the generated images and the reference images, referred to as CLIP-I and DINO, respectively. 
To evaluate multi-subject generation, we extended CLIP-I and DINO by computing the average similarity between each generated image and all corresponding reference images. 
For text fidelity, we used the CLIP model to calculate the cosine similarity between the generated images and the text prompts, which is known as CLIP-T.
To ensure statistical reliability, each test sample was generated with four different random seeds.

\begin{figure}[ht]
    \centering
    \includegraphics[width=1.0\textwidth]{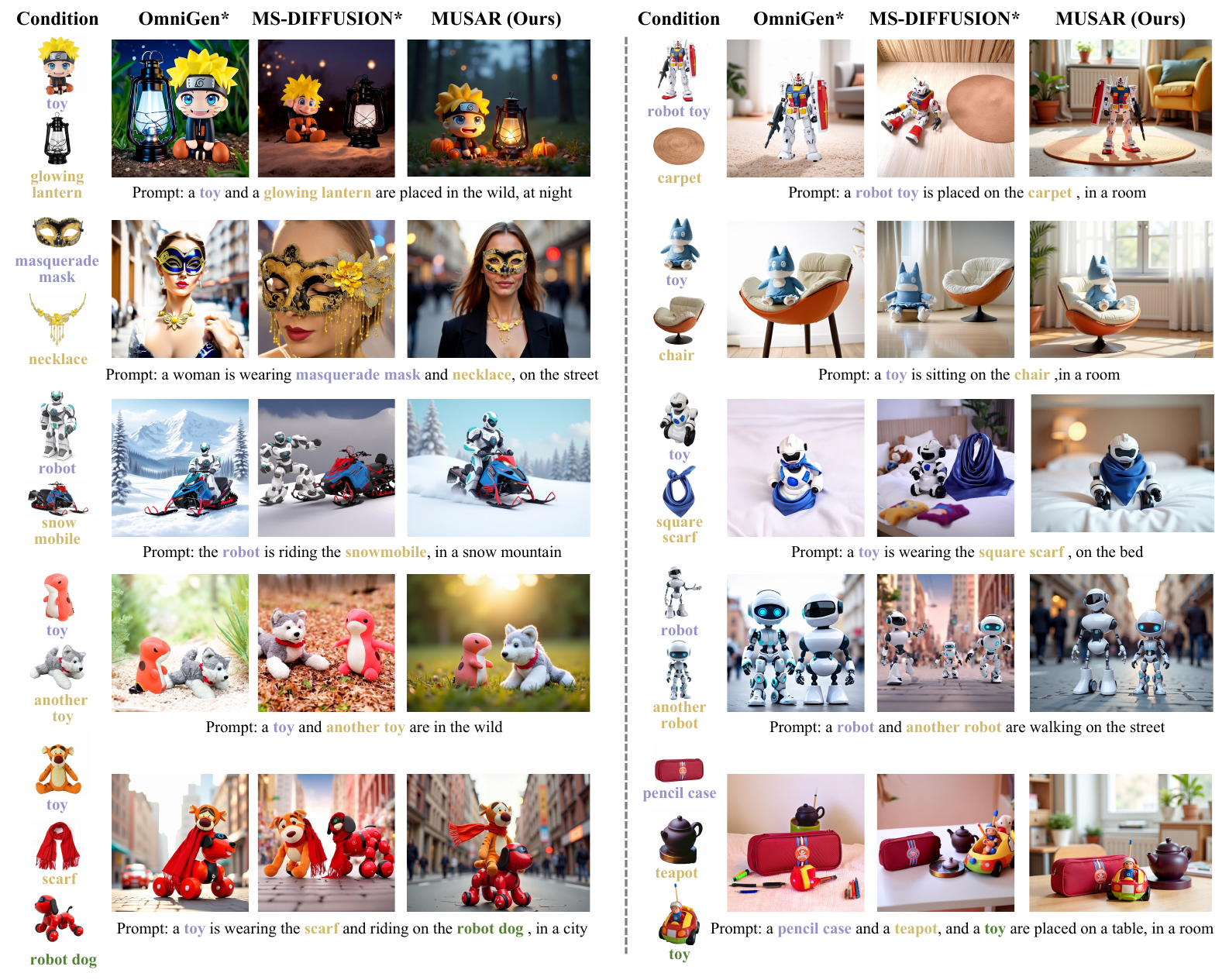}
    \caption{Qualitative comparison with several state-of-the-art methods on multi-subject customization.  $^{*}$ denotes the method training on multi-subject dataset.}
    \label{fig:ms_qualitative}
\end{figure}

\begin{figure}[ht]
    \centering
    \includegraphics[width=1\textwidth]{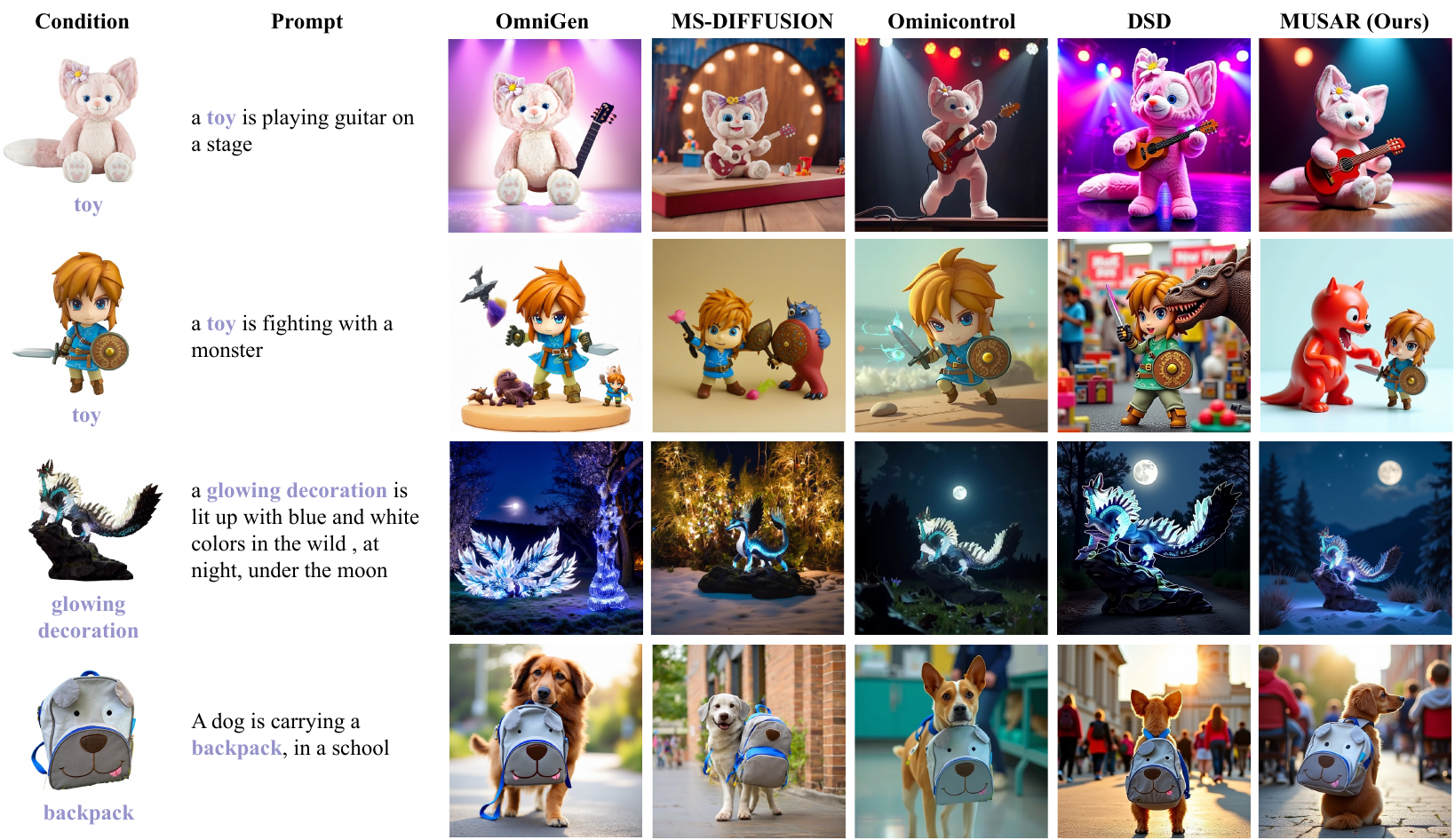}
    \caption{Qualitative comparison on single-subject customization.}
    \label{fig:ss_qualitative}
\end{figure}

\subsection{Qualitative Comparison}
We conduct comprehensive qualitative comparison with state-of-the-art multi-subject customization methods, including Omnigen ~\cite{xiao2024omnigen} and MS-Diffusion ~\cite{wang2025ms}, as shown in Figure ~\ref{fig:ms_qualitative}.
It is worth to note that while all comparison methods are trained on carefully curated multi-subject datasets, our MURSAR achieves superior performance in subject consistency, attribute disentanglement, and visual fidelity using only single-subject training data.
Specifically, for the case of two subjects (row 1-4), our method generates naturally coordinated compositions, in contrast to baseline methods that often produce artificial copy-paste artifacts or physically implausible interactions.
For instance, as shown in Figure~\ref{fig:ms_qualitative} our method accurately renders the lighting interaction between the lantern and figurine (first row, left), demonstrates plausible sitting postures (second row, right), and correctly positions the scarf around the robot’s neck (third row, right).
Despite being trained exclusively on two-subject data, MUSAR shows remarkable generalization to multi-object scenarios (row 5), surpassing all competing methods. This capability stems from our novel Dynamic Attention Routing mechanism, which intelligently allocates optimal conditions to each spatial regions, effectively mitigating multi-subject interference.

Furthermore, we conducted qualitative comparisons with single-subject methods, including Omnigen ~\cite{xiao2024omnigen} and MS-Diffusion ~\cite{wang2025ms}, OminiControl and DSD ~\cite{cai2024diffusion}. The results demonstrate
that while our method primarily focuses on multi-subject preservation, it simultaneously achieves state-of-the-art performance in single-subject customization .

\subsection{Quantitative Comparison}

As demonstrated in Table~\ref{tab:quant}, our method outperforms existing approaches in both single-subject and multi-subject customization tasks. 
As one can see, MUSAR achieves the highest scores across four metrics compared to all baseline methods.
Particularly noteworthy is MUSAR's performance in multi-subject scenarios: despite being trained solely on single-subject dataset, it surpasses the method (OmniGen~\cite{xiao2024omnigen}, MS-DIFFUSION~\cite{wang2025ms}) using specialized multi-subject data in visual fidelity metrics (DINO and CLIP-I) while maintaining comparable text alignment (CLIP-T). 
These results clearly demonstrate MUSAR's superior performance in preserving subject identity and attributes across different customization scenarios.

\setlength{\tabcolsep}{2.4mm}{
\begin{table*}[t]
\caption{Quantitative comparisons of single-subject and multi-subject customization on DreamBench. $^{*}$ denotes the method training on multi-subject dataset. The best results are shown in bold.}
\begin{tabular}{c|ccc|ccc}
\Xhline{1pt}
{\multirow{2}{*}{Mothod}} & \multicolumn{3}{c|}{Single-subject Customization}    & \multicolumn{3}{c}{Multi-subject Customization}      \\ \Xcline{2-7}{0.5pt}
                      & DINO$\uparrow$ & CLIP-T$\uparrow$ & CLIP-I$\uparrow$ & DINO$\uparrow$ & CLIP-T$\uparrow$ & CLIP-I$\uparrow$ \\ \Xhline{0.5pt}
OminiControl~\cite{tan2024ominicontrol} & 0.720 & 31.07 & 0.804 & -- & -- & --\\
DSD ~\cite{cai2024diffusion} & 0.752 & 31.06 & 0.811 & -- & -- & --  \\
OmniGen$^{*}$ ~\cite{xiao2024omnigen}  & 0.765 & 31.01 & 0.820 & 0.691 & 33.17 & 0.716\\
MS-DIFFUSION$^{*}$ ~\cite{wang2025ms}& 0.735 & \textbf{31.91} & 0.819 & 0.678 & \textbf{34.20} & 0.711
\\ \Xhline{0.5pt}
\textbf{MUSAR (Ours)} & \textbf{0.774}  & 30.29 & \textbf{0.833} & \textbf{0.704} & 33.90 & \textbf{0.720}                                                                          \\ \Xhline{1pt}
\end{tabular}
\label{tab:quant}
\end{table*}
}

\subsection{Ablation Study}
We conduct comprehensive ablation experiments to analyze the impact of each model component. The experimental setup involves removing the following elements from our full model.

\textbf{w/o Diptych data.} This model is training without the constructed diptych data and relies exclusively on single-subject data for training.

\textbf{w/o Diptych Biases Mitigation.} This model removes the Diptych Biases Mitigation: Static Attention Routing module and Dual-branch LoRA, enabling prompt-condition and inter-condition interactions, uniformly use a single set of LoRA to fine-tune all parameters.

\textbf{w/o Dynamic Attention Routing. }This model removes the Dynamic Attention Routing module, enabling each region to simultaneously attend to multiple subjects.

\textbf{Full Model. } De-biased diptych learning and Dynamic Attention Routing are applied in this model.

\begin{figure}[!t]
    \centering
    \includegraphics[width=1.0\textwidth]{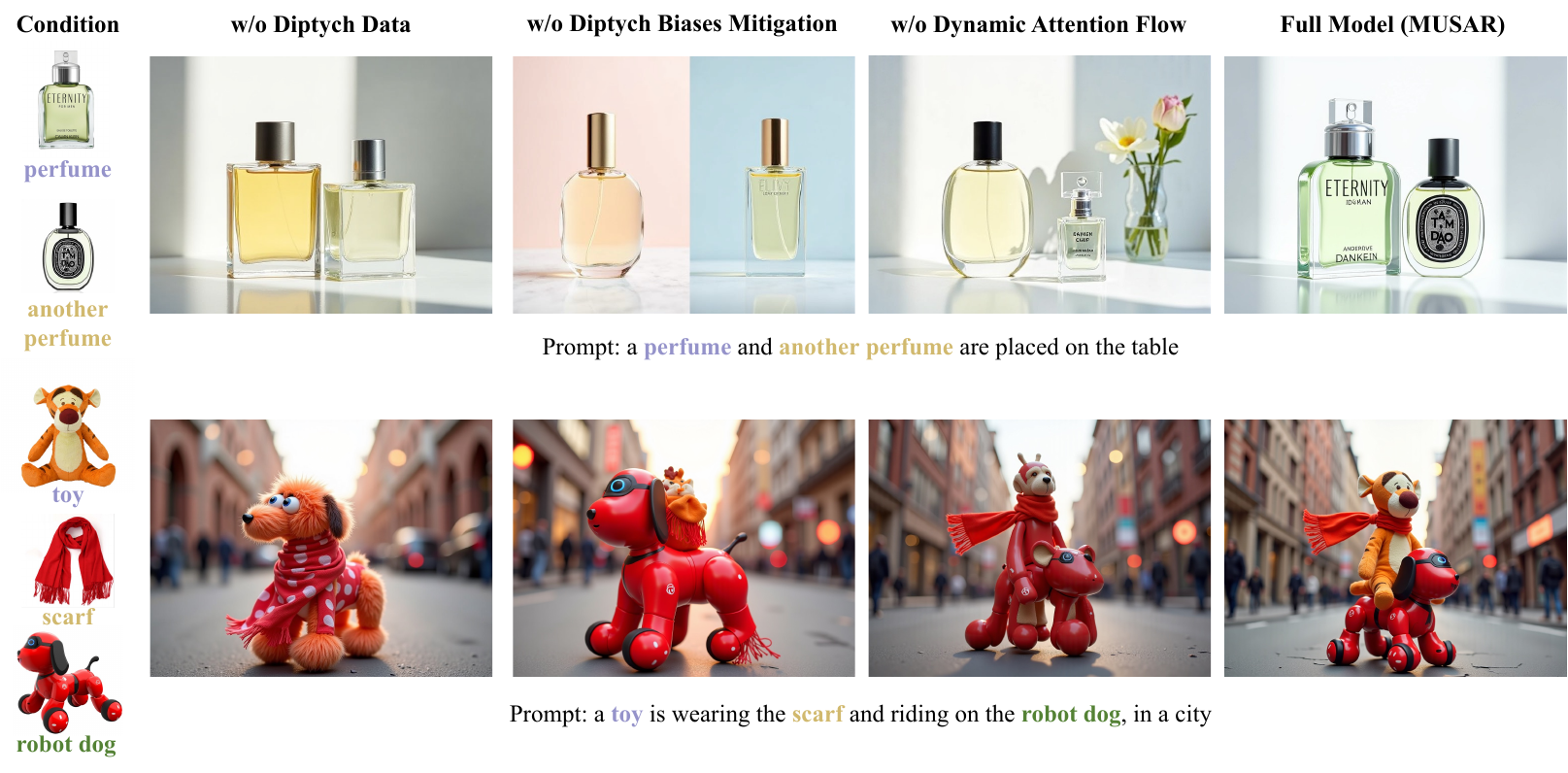}
    \caption{Qualitative comparison of the ablation study.}
    \label{fig:ablation}
\end{figure}

Figure ~\ref{fig:ablation} presents the qualitative results of our ablation study. For models without diptych learning (column 1),  the generated samples exhibit poor multi-subject consistency. This conclusively demonstrates the importance of diptych learning for preserving multi-subject characteristic. For models without static attention flow (column 2),
the model tend to learn biases from the diptych data prompts during training, leading to generate diptych results during inference, and causes the loss of elements included in the prompt. For example, the case in row 2, the model erroneously merges multiple subjects in the generated output, failing to properly respond to the discrete objects specified in the text prompt.
For models without dynamic attention flow (column 3), the model exhibits significant cross-subject confusion, manifesting in erroneous attribute entanglement between distinct subjects - as exemplified in row 2 by the toy incorrectly adopting the robotic dog's color.
Thanks to our carefully designed de-biased diptych learning and dynamic attention routing mechanism, our proposed MUSAR demonstrates the remarkable capability of learning complex multi-object customization solely from single-subject datasets.

\section{Conclusion}\label{sec:conclusion}

We present MUSAR, a novel framework for multi-subject customization that learns effectively from single-subject data. To address data limitations, we propose debiased diptych learning, which synthesizes diptych training pairs from individual subject images while correcting systemic bias through static attention routing and dual-branch LoRA adaptation. For cross-subject entanglement, we develop dynamic attention routing that employs spatial gating to guide image regions to associate with their corresponding subjects. Quantitative and qualitative results comprehensively demonstrate MUSAR's superiority over state-of-the-art methods across image fidelity, subject consistency, and interaction naturalness, while requiring only single-subject training dataset.

\bibliographystyle{plain}
\bibliography{ref}

\end{document}